\documentclass[journal]{IEEEtran}

\usepackage{cite}
\usepackage[pdftex]{graphicx}

\begin{document}

\title{A Compact Two-Phase Twisted String Actuation System: Modeling and Validation}
%
%
%

\author{Mahmoud Tavakoli$^{1}$, Rafael Batista$^{1}$ and Pedro Neto$^{2}$%

\thanks{$^{1}$Institute for Systems and Robotics,
        Dep. of Electrical and Computer Eng.,
        University of Coimbra, Portugal.
        {\tt\footnotesize \{rafaeljcb,mahmoud\}@isr.uc.pt}
		  $^{2}$CEMUC, Dep. of Mechanical Eng., University of Coimbra
		{\tt\footnotesize pedro.neto@dem.uc.pt}}
}

\maketitle

\begin{abstract}
\textbf{In this paper, we propose a compact twisted string actuation system that achieves a high contraction percentage (81\%) on two phases: multi string twist and overtwist. This type of system can be used in many robotic applications, such as robotic hands and exoskeletons. The overtwist phase enables the development of more compact actuators based on the twisted string systems. Furthermore, by analyzing the previously developed mathematical models, we found out that a constant radius model should be applied for the overtwisting phase. Moreover, we propose an improvement of an existing model for prediction of the radius of the multi string system after they twist around each other. This model helps to better estimate the bundle diameter which results in a more precise mathematical model for multi string systems. The model was validated by performing experiments with 2, 4, 6 and 8 string systems. Finally, we performed extensive life cycle tests with different loads and contractions to find out the expected life of the system.}
\end{abstract}


%
\IEEEpeerreviewmaketitle

\section{Introduction}
%
%
%
%

\label{sec:INTRODUCTION}

A twisted string actuation system converts the rotation from a motor's shaft into a linear motion. It offers some advantages over conventional systems, like a rack and pinion, or a lead screw and nut system, such as being lighter, simpler and less expensive. The actuator is also not completely rigid due to the strings elastic behavior. This can make it harder to control but it should be safer to operate around humans, due to the natural compliance of the entire system.\\

Some of the earliest applications of the twisted string systems in robotics was reported by M. Suzuki et. al., where they were used to drive a six-legged robot\cite{Strand-MuscleOctober2007} and also an articulated arm that includes an anthropomorphic robotic hand \cite{Strand-Muscle2004,Strand-Muscle2007}. As reported in \cite{Strand-Muscle2007}, actuators of the anthropomorphic hand are placed in the forearm of the system outside the hand itself. The same concept was applied in the DEXMART hand \cite{DEXMARTJuly2013}, where several actuators with twisted string systems are placed outside the palm and in a relatively large forearm. While placing the actuators at a remote distance from the driven object is advantageous for some applications, it is not desired in many others. Godler et. al. presented a five fingered robotic hand, that uses 15 twist drives to control the fingers \cite{TwistDrive2010}. In this case the actuators are placed in the fingers. To do so and in order to increase the joints motion range with a limited stroke of the twisted string system, the coupling point of the tendon was shifted from the joint at the cost of making the fingers larger.\\

All above examples demonstrate a common problem in the twisted string systems. That is, twisted string systems suffer from a low contraction to length ratio (the ratio of the actuator stroke to the total string length). If neglecting the dead zone (actuators and coupling zone), this ratio for a lead screw and nut system can reach to almost 100 \%, which is the same for a rack and pinion system. Previous works on twisted string systems,\cite{TwistDriveNov2011, TwistedStringActuationSystemApril2013, TwistedStringExoOctober2012, TwistedStringExoApril2013}, show a contraction percentage of less than 25\%. This is problematic for systems that demand for compactness. For instance, one application of the twisted string system, which was considered by several research teams, is in robotic hands. For this application, it is desirable to develop light weight self-contained hands with all actuators integrated inside the palm.\\

\subsection{The two phase twisted string system}

In this paper, we propose a two phase twisted string concept which could achieve a much higher contraction compared to the previous systems. In this concept, in the first phase of the contraction, multiple strings twist around each other to the maximum possible amount. At the end of this phase, a single twisted string is formed, in which the individual strings cannot twist around each other anymore. Afterwards, in the second phase, the single string twists around itself. We call this stage ``overtwisting phase''. In this way, we could achieve a significant contraction percentage of 81\% which was never reported before.\\

\begin{figure}[tbph]
      \centering
      \includegraphics[trim =30 170 170 460, clip=true, width=110 mm]{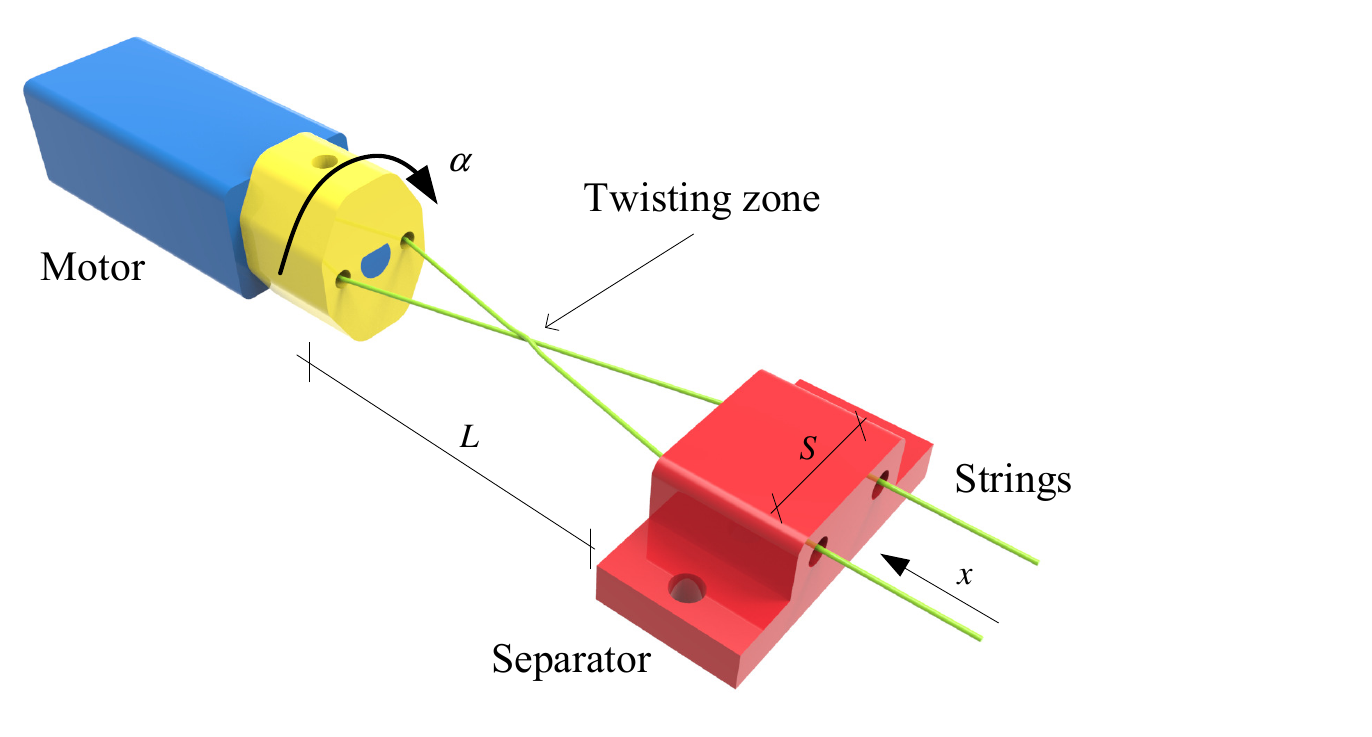}
      \caption{\textbf{Twisted string system: electric motor, strings, connection between motor shaft and strings and separator. $L$ is twisting zone length, $\alpha$ is the rotational angle of the motor shaft, $x$ is the linear displacement of the strings and $S$ is the distance between the holes of the separator.}}
      \label{fig:systemSchematics}
\end{figure}

Figure \ref{fig:systemSchematics} shows the schematic of our proposed twisted string system. It is composed by the actuator, a connection between the actuator shaft and the strings and strings. 
Similar to the mechanism presented by Popov et. al.  \cite{TwistedStringExoNovember2013}, we also used a separator. The role of the separator is to limit the zone where the strings can twist. In this way, beyond the separator there is only linear motion. The zone between the shaft and the separator (Figure \ref{fig:systemSchematics}) is called the twisting zone.\\

In all the above articles, the authors either did not specify the reason for their limited contraction range or mentioned that the tensile strength and fast wearing of the strings are their limiting factor. For instance in \cite{TwistedStringExoOctober2012}, the authors stated that most strings suitable to be used in twisted string can only carry relative small loads.
Furthermore, the overtwisting stage was not considered in any of the previous works, \textbf{while our proposed system was tested for the overtwisting phase for hundred of cycles and performed a life cycle experiment that demonstrated that the strings can endure overtwisting without a reduction in the strings life (Section \ref{sec:LifeCycle}).}\\

\subsection{Mathematical model}

In this article we suggest a mathematical model for controlling the system linear displacement over the whole range of the contraction. Even though previous mathematical models could predict the contraction in the initial twisting zone, \textbf{none of them could correctly} fit our experimental results after 15\% of the contraction.
In the twisted string system presented in \cite{TwistedStringExoOctober2012}, authors mentioned that their mathematical model provides high correlation with the practical data only up to 15\% contraction.\\

After studying the previous models and our experiments, we performed some corrections to the model presented in \cite{TwistedStringExoNovember2013}, which applies to systems with a separator. First, \textbf{it was} developed an improved theory for prediction of the radius of the multi string system after they twist around each other. Second, we showed that a constant radius model performs significantly better over the variable radius model for the long stroke range. \textbf{Integrating these two contributions, this paper suggests} a mathematical model that can precisely predict the contraction of the system over a large range of contraction.\\

\section{EXPERIMENTAL SETUP}
\label{sec:EXPERIMENTAL}

\begin{figure}[tbph]
      \centering
      \includegraphics[trim =15 420 120 110, clip=true, width=125 mm]{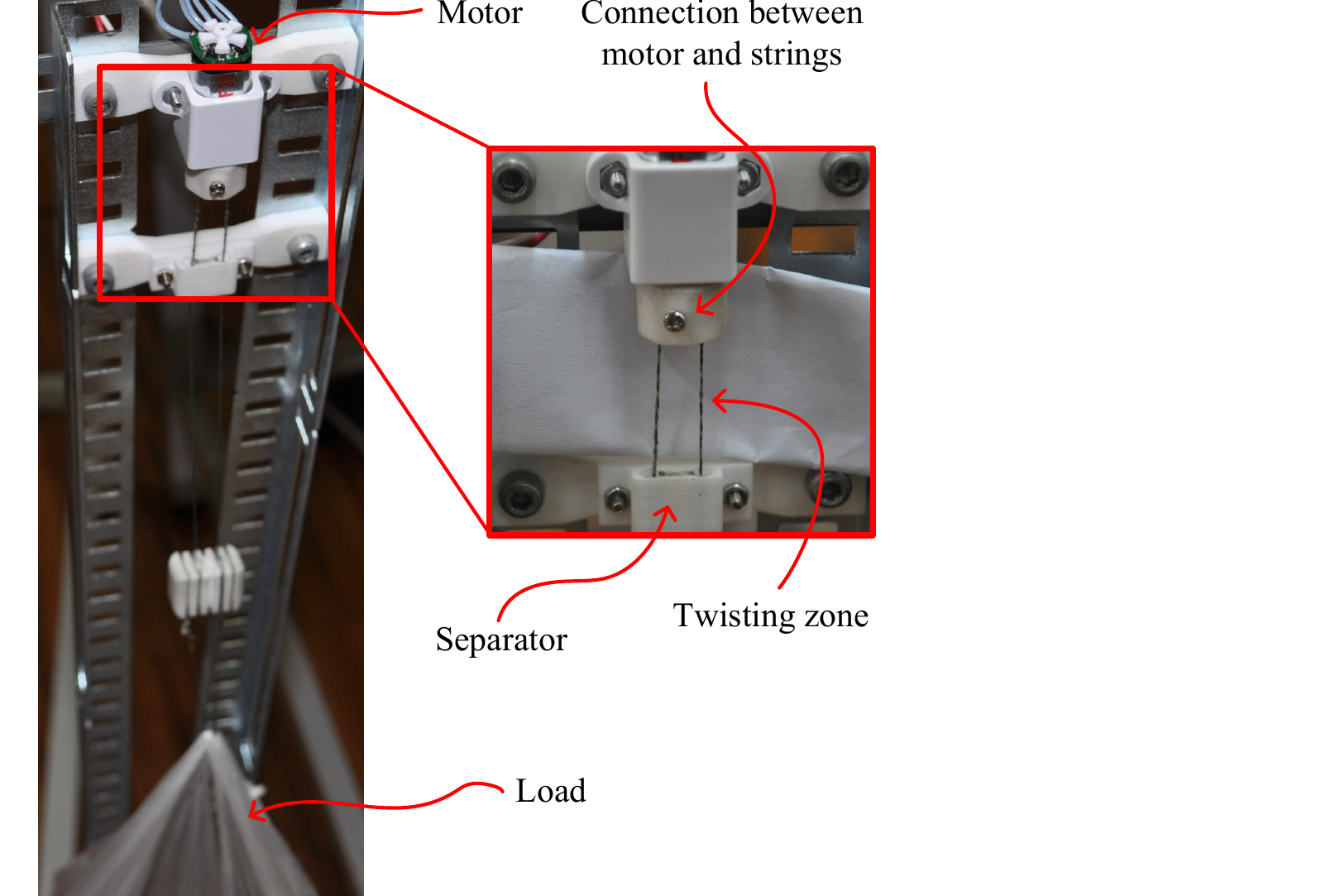}	   
      \caption{Twisted strings actuation system prototype that was used to lift a 2 kg weight.}
      \label{fig:firstPrototype}
\end{figure}

In order to test the two phase twisting string system a prototype was built (figure \ref{fig:firstPrototype}). \textbf{It was used a Micro Metal Gearmotor from Pololu with 100:1 gear ratio, 320 r.p.m. and 0.22 Nm stall torque.} The string which was used for this experiment, was a "Potenza Braided Line" from Vega, with a diameter of 0.35 mm, \textbf{that according to the manufacturer is capable of enduring loads up to 321 N} and has a high resistance to wearing. The twisted string actuation system was tested by lifting 2 kg weights with 2, 4, 6 and 8 strings and the linear displacement for each turn of the motor was recorded. The dimensions of the system are shown in the Table \ref{table:systemDimensions}.\\

\begin{table}[tbph]
\caption{Dimensions of the twisted string actuation system.}
\label{table:systemDimensions}
\renewcommand{\arraystretch}{1.2}
\centering
\begin{tabular}{|c|c|c|c|}
\hline
Nr.  & L, Twisting zone & S, Distance between & Diameter$^1$ \\
of strings & length [mm] & the separator holes [mm] & [mm] \\
\hline
2 & 23.20 & 5 & 0.47\\
\hline
4 & 23.42 & 5 & 0.70\\
\hline
6 & 22.85 & 5 & 0.86\\
\hline
8 & 23.30 & 5 & 0.99\\
\hline
\multicolumn{4}{l}{$^1$Diameter of the string bundle after 5 turns.} \\
\end{tabular}
\end{table}

\begin{figure}[tbph]
	\centering
	\includegraphics[trim =100 130 120 150, clip=true, width=90 mm]{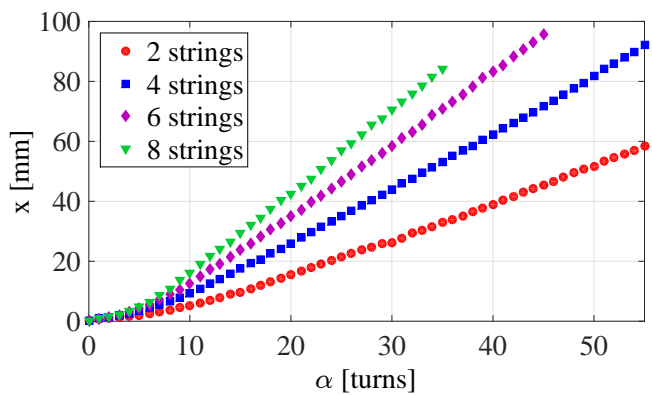}	   
	\caption{\textbf{Experimental results for various number of strings in the system, with $x$ being the linear displacement and $\alpha$ the number of turns at the output shaft of the gearmotor.}}
	\label{fig:graphExperimentalResults}
\end{figure}

\begin{figure*}[!t]
      \centering
      \includegraphics[trim =0 5 0 20, clip=true, width=180 mm]{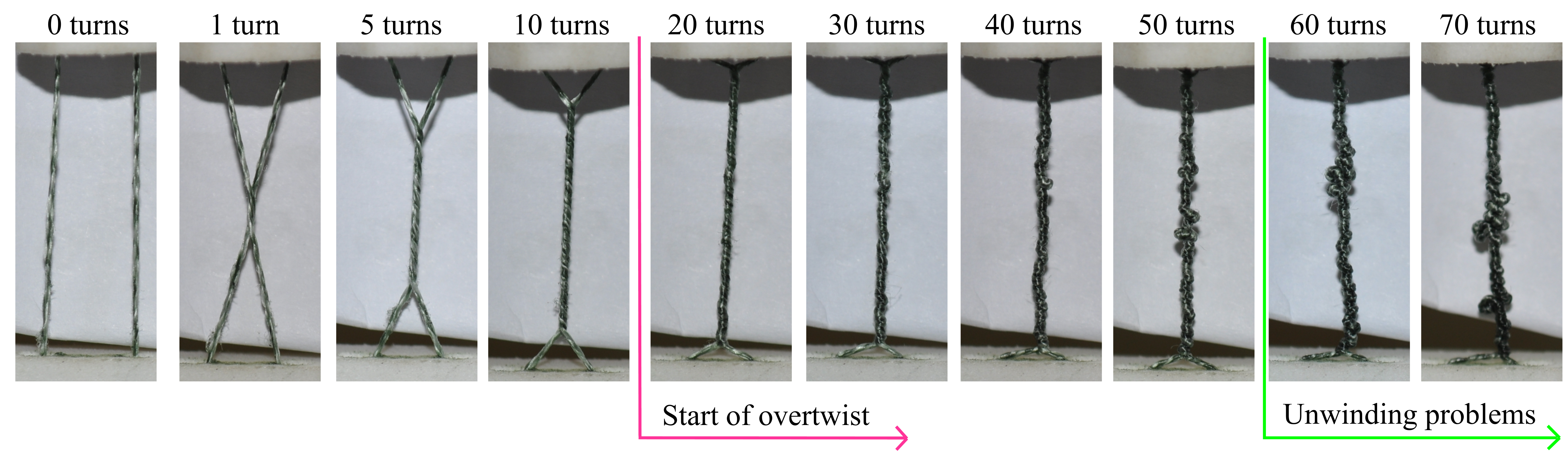}	   
      \caption{Snapshot of the twisted string system with 2 strings from 0 to 70 turns. As can be seen, overtwisting starts to happen after 20 turns. After 40 turns, overtwist is not evenly distributed over the twisting zone and local overtwist starts to happen, which can be seen in 50 turns snapshot.}
      \label{fig:strings}
\end{figure*}

Figure \ref{fig:graphExperimentalResults} shows the experimental results where the linear displacement is compared against the number of turns of the output shaft. By increasing the number of strings, the diameter of the string bundle also increases. As can be seen in Figure \ref{fig:graphExperimentalResults}, at the same rotation input, higher displacements are possible with higher number of strings, due to the increase in diameter. Untwisting problems occur because the knots that form after several rounds of overtwisting (Figure \ref{fig:strings} after 60 turns), get stuck in the separator holes when the system is unwinding. In Figure \ref{fig:graphExperimentalResults} it is possible to observe that by increasing the number of strings, untwisting problems happened at an earlier stage (less number of turns). Table \ref{table:Contraction} shows the amount of contraction achieved by the twisted string system for a different number of strings. It is important to note that in this specific case, \textbf{the system with 6 strings is the one that allows to achieve the most significant linear displacement}, $x$, \textbf{and naturally the biggest contraction percentage} (81\%). Since the twisting zone length is constant, when the system contracts, \textbf{more strings are inside the twisting zone}. Then the total string length is calculated as the sum of the length of the twisting zone with the maximum contraction achieved. The total length of the system can be bigger in the experiment. However, in the calculation of the contraction ratio we only consider the effective length of the system. This is due to the fact that the strings that do not enter the twisting zone have little effect on the behavior of the system, and therefore are not accounted for.\\

\begin{table}[tbph]
\caption{Achieved contraction}
\label{table:Contraction}
\centering
\begin{tabular}{|c|c|c|c|c|}
\hline
Nr. of & Twisting zone & Contraction & Total string & Contraction\\
strings & length [mm] & [mm] & length$^1$ [mm] & [\%]\\
\hline
2 & 23.20 & 58.40 & 81.60 & 72\\
\hline
4 & 23.42 & 92.02 & 115.44 & 80\\
\hline
6 & 22.85 & 95.67 & 118.52 & 81\\
\hline
8 & 23.30 & 84.09 & 107.39 & 78\\
\hline
\multicolumn{5}{l}{$^1$The total string length corresponds to the sum of the twisting zone}\\
\multicolumn{5}{l}{length with the contraction.}\\
\end{tabular}
\end{table}

\subsection{Comparison between different separators and motor shaft connectors}

\begin{figure}[tbph]
      \centering
      \includegraphics[trim =400 10 320 80, clip=true, width=90 mm]{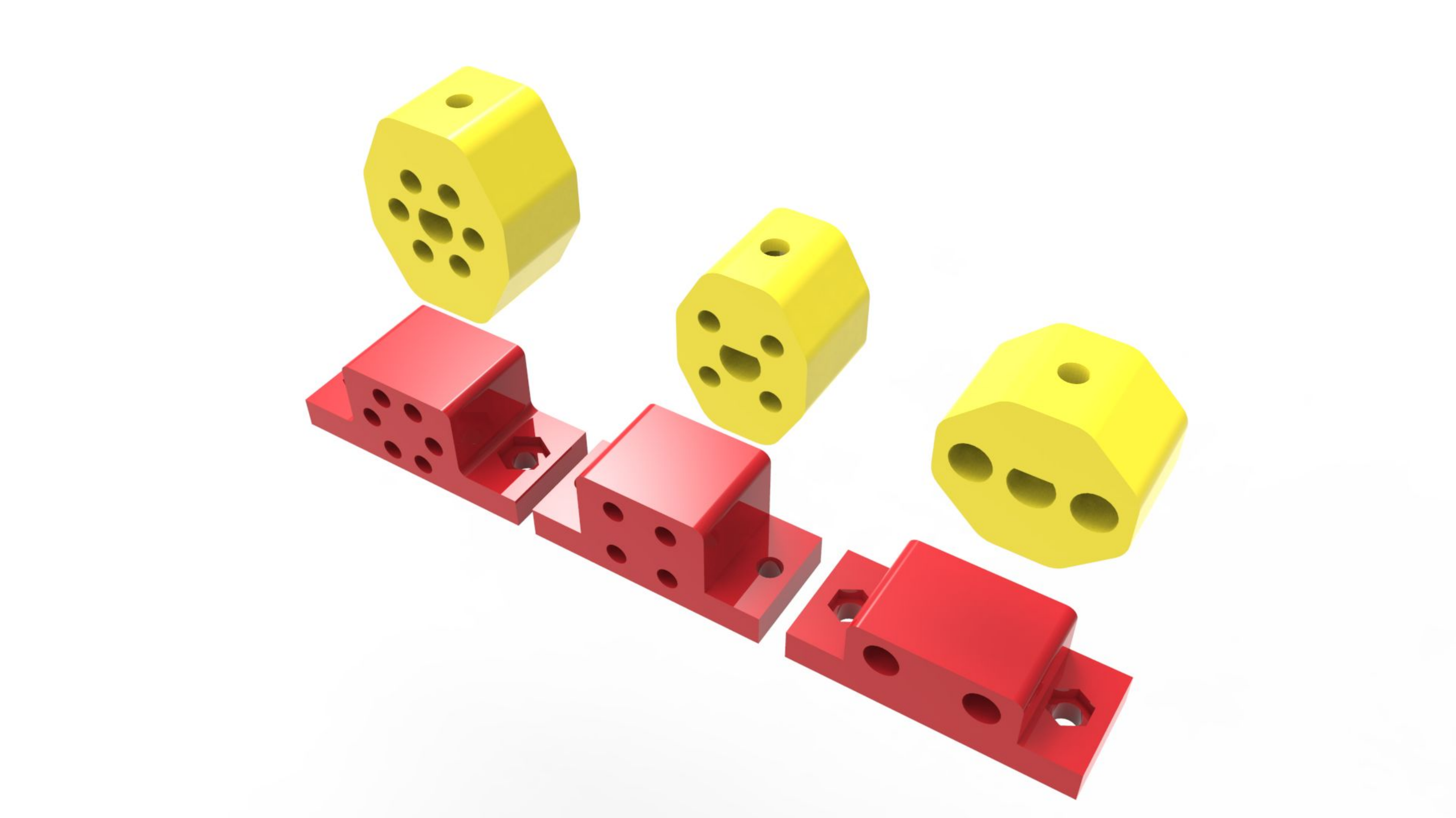}	   
      \caption{The different separators and motor shaft connectors tested. The distance between the holes of the separator, $S$, remains the same in all of them.}
      \label{fig:separators}
\end{figure}

In order to be able to increase the diameter of the string bundle, more strings need to be added. The strings can simply be passed through the two holes system, or more holes need to be added to accommodate all of the strings. To find if there is any difference in behavior due to the number of  holes, both solutions were tested, Figure \ref{fig:separators}.\\

\begin{figure}[tbph]
      \centering
      \includegraphics[trim =30 240 25 255, clip=true, width=90 mm]{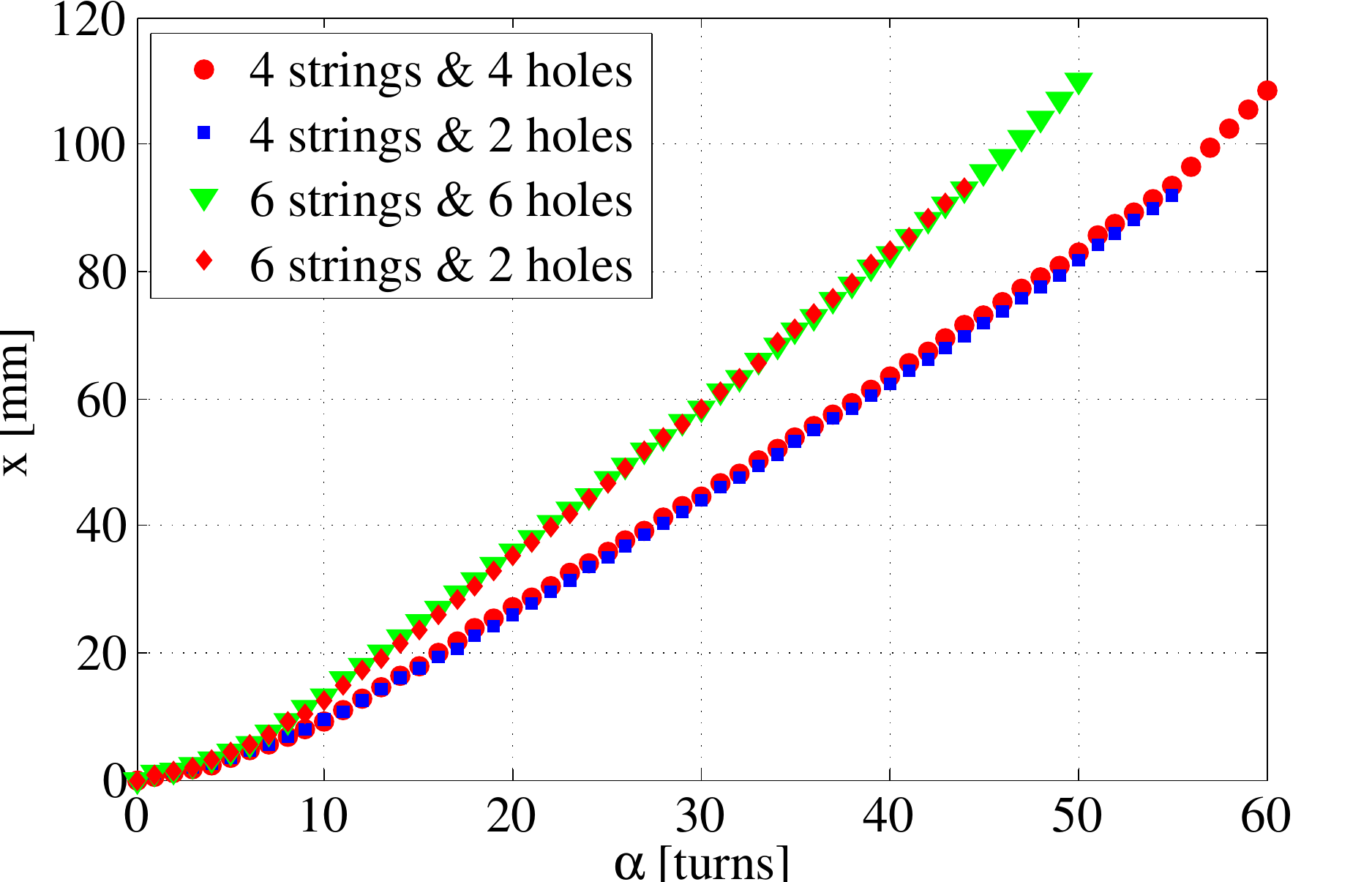}	   
      \caption{Comparison of the variation of the linear displacement, $x$, with the number of turns at the output shaft of the gearmotor, $\alpha$, for the various separators and connectors tested.}
      \label{fig:graphSeparators}
\end{figure}

Figure \ref{fig:graphSeparators} shows the variation of linear displacements with the rotation of the motor shaft. Results show that the number of holes has no influence on the displacement of the system. However, the systems with 4 and 6 holes had much less untwisting problems than the 2 holes system.\\

\subsection{Twisting and Overtwisting}

Figure \ref{fig:strings} demonstrates several snapshots of the twisted string system with 2 strings from 0 to 70 turns. As can be seen, overtwisting starts to happen after 20 turns. After 40 turns, overtwist is not evenly distributed over the twisting zone and local overtwists start to happen, which can be seen in 50 turns snapshot. After this stage, the string is not uniform anymore. Such local overtwists move along the strings in the twist zone as can be seen in 60 and 70 turns snapshots.\\

Table \ref{table:Overtwist} shows the amount of contraction and the number of turns where overtwist began for various number of strings. This is the result of observation during the experiments. The overtwist phase seems to start between 10\% and 20\% of contraction, which is about the maximum contraction percentage that similar twisted string systems have reported. \textbf{The results presented in table \ref{table:Overtwist} demonstrate that the increase in diameter makes the overtwist phase to occur in less turns.} \textbf{Since the length of the twisting zone is constant, and then the volume of the strings inside the twisting zone increases with the additional strings.} 

\begin{table}[tbph]
\caption{Start of overtwist for various number of strings.}
\label{table:Overtwist}
\renewcommand{\arraystretch}{1.2}
\centering
\begin{tabular}{|c|c|c|c|}
\hline
Nr. of strings & Diameter [mm] & Contraction [\%] & Overtwist [turns]\\
\hline
2 & 0.47 & 19 & 20\\
\hline
4 & 0.70 & 9 & 11\\
\hline
6 & 0.86 & 10 & 10\\
\hline
8 & 0.99 & 10 & 8\\
\hline
\end{tabular}
\end{table}

\subsection{Maximum Contraction}
It is important to understand what limits the maximum contraction in this system.
Experiments revealed that what limits the contraction is not how much the system can overtwist, but if that amount of overtwisting is reversible.\\

\begin{figure}[tbph]
      \centering
      \includegraphics[width=30 mm]{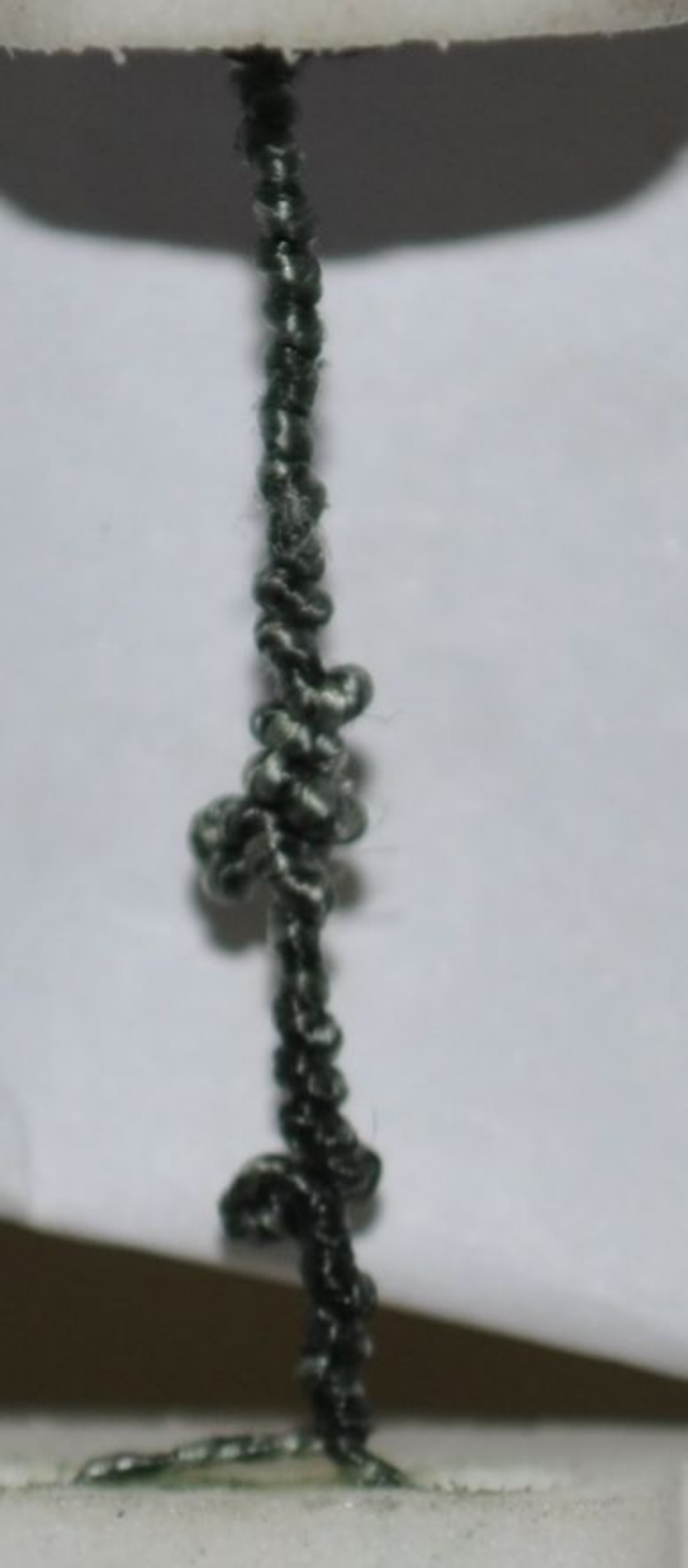}	   
      \caption{Zoom of strings when they are twisted for 70 turns of the output shaft of the gearmotor.}
      \label{fig:StringsZoom70}
\end{figure}

In figure \ref{fig:strings} the overtwisting is noticeable after 20 turns. After 50 turns it is possible to observe formation of local overwisting zones with larger diameters than the rest of the strings. This may form some knots in the strings that cause problems unwinding, because the large knots get stuck in the separator holes and the only way to untwist the system is by manually help the strings unwind. Figure \ref{fig:StringsZoom70} shows a close snapshot of the system after 70 turns where untwisting is not reversible. We observed that these non reversible knots always happened at the same number of turns.\\

In table \ref{table:lenghtVSdisplacement} the contraction of the system reported here is compared with other proposed systems. The length present in table \ref{table:lenghtVSdisplacement} is the sum of the length of the twisting zone and the maximum linear displacement for systems with a separator. For systems without a separator the total length of strings is considered. This means that the length of the actuator is not taken into account in this comparison.\textbf{ According to table \ref{table:lenghtVSdisplacement} the system proposed clearly outperforms other systems in terms of contraction percentage.} The highest contraction percentage was possible with a 6 string system that reached 96 mm of linear displacement in the total length of 119 mm, which corresponds to 81\% of contraction, compared to maximum 24\% in previous systems. This demonstrates that the overtwisting phase can significantly improve the compactness of such systems.

\begin{table}[tbph]
\caption{Comparison between length of the system and linear displacement}
\label{table:lenghtVSdisplacement}
\renewcommand{\arraystretch}{1.2}
\centering
\begin{tabular}{|c|c|c|c|}
\hline
Previous & Length of & Maximum linear & Contraction\\
works & the system [mm] & displacement [mm] & [\%]\\
\hline
\cite{TwistDriveNov2011} & 55 & 13 & 24\\
\hline
\cite{TwistedStringActuationSystemApril2013} & 200 & 28 & 14\\
\hline
\cite{TwistedStringExoOctober2012} & 235 & 28 & 12\\
\hline
\cite{TwistedStringExoApril2013} & 250 & 60 & 24\\
\hline
Our system & 119 & 96 & 81\\
\hline
\end{tabular}
\end{table}

\section{MATHEMATICAL MODEL}
\label{sec:MATHEMATICAL MODEL} 

In order to be able to control the tendon based mechanisms with a twisted string system, we need to estimate the contraction of the tendon, based on the position of the rotary actuator's shaft. None of the previously developed models could correctly fit our experimental results. After studying the previous models and according to our experiments, we performed some adjustments to the models. The model presented in \cite{TwistedStringExoNovember2013}, applies to systems with a separator and can predict the system displacement with a good precision for small contractions. However, this model starts to develop increasing errors for larger contraction amounts. \textbf{In the aforementioned model}, despite the existence of a separator, the distance between the holes of the separator was not counted in the formulation. In another research work \cite{TwistDrive2010}, in a system that does not contain a separator, researchers considered the offset between the string attachment points in their mathematical model. Inspired by this work, we applied this offset in the mathematical formulation as the distance between the separator holes.
Then, we developed an improved theory for prediction of the radius of the multi string system after they twist around each other. Finally, we showed that a constant radius model performs significantly better over the variable radius model for the long stroke range. 
As it will be shown in the results, the mathematical model presented here,  can predict the system displacement for large contraction ratios of up to 81\%.\\

The influence of the force applied was not considered in the prediction of the linear displacement. In \cite{TwistedStringExoOctober2012}, the authors stated that the change load force has very little influence in the mathematical model of the twisted string system. We chose 20, 30 and 50 N loads to test our system, to be able to compare our results with similar works like \cite{TwistedStringExoAug2014} that used loads up to 31.5 N.\\

\subsection{Mathematical model for systems with a separator}

While the architecture of our system is similar to the system presented in \cite{TwistedStringExoApril2013, TwistedStringExoNovember2013}, we observed that the mathematical model presented in the articles can only predict the contraction up to 15\% of the original length. 
One of the parameters that was not considered in \cite{TwistedStringExoApril2013, TwistedStringExoNovember2013} is $S$, the distance between the holes in the separator. In \cite{TwistDrive2010}, authors considered this distance in their formulation, however their model does not apply for systems with a separator.\\

\begin{figure}[tbph]
      \centering
      \includegraphics[trim =150 90 130 520, clip=true, width=90 mm]{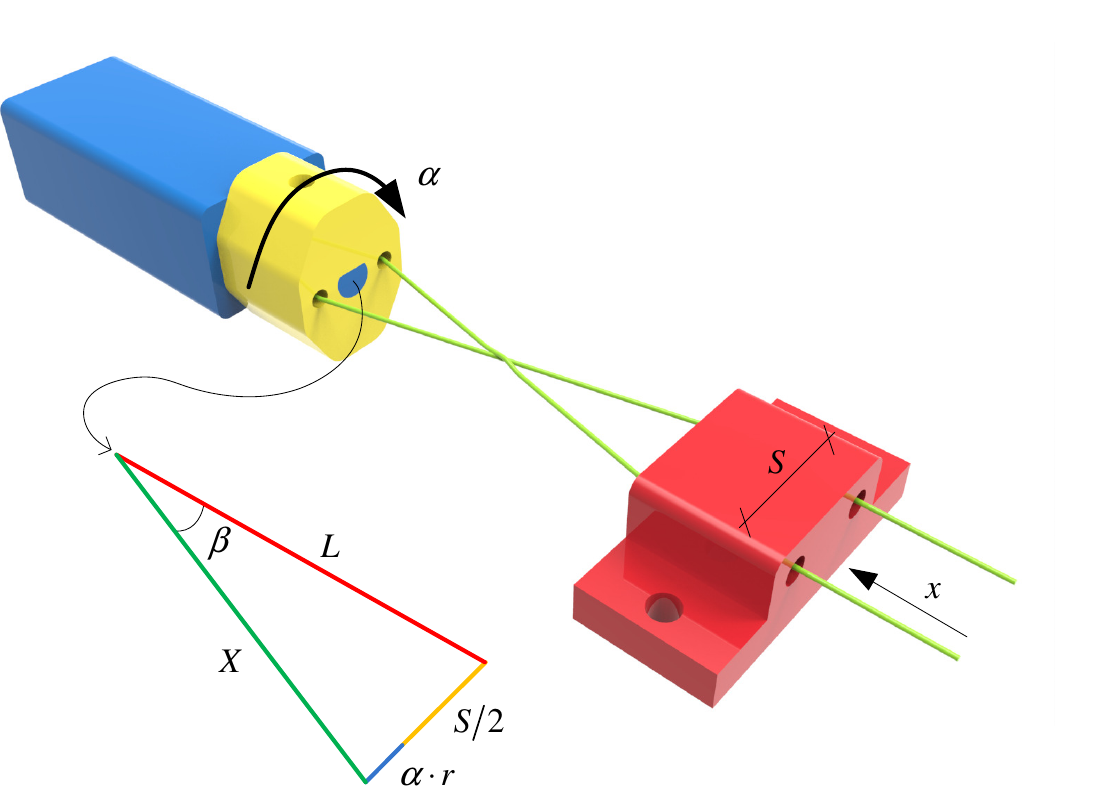}
      \caption{\textbf{String schematics and proposed model.}}
      \label{fig:stringsSchematics}
\end{figure}

Figure \ref{fig:stringsSchematics} shows the schematic of the proposed system that includes the $S$ parameter. Here $L$ is the length of the twisting zone, $\alpha$ is the rotation angle of the motor shaft, $r$ is the radius of the string bundle, $X$ is the length of the strings inside the twisting zone and $\beta$ is helix angle.\\

When the motor shaft rotates, only $X$, $\alpha$ and $\beta$ will increase while the parameters $L$, $r$ and $S$ remain constant. If the length of the string inside the twisting zone, $X$, increases with the rotation of the motor shaft, then the linear displacement, $x$, is equal to the amount of string that enters the twisting zone. \textbf{The proposed method assumes the same model proposed in \cite{TwistDrive2010}, in which the vertex of the triangle that forms angle $\beta$ is coincident with the centre of the motor, Figure \ref{fig:stringsSchematics}.} Therefore, according to \cite{TwistedStringExoApril2013, TwistedStringExoNovember2013}, the following equations can be used to calculate the linear displacement:
 
\begin{equation}
x=X-L
\label{eq:difference}
\end{equation}

\begin{equation}
x=\sqrt{L^2+\left(\frac{S}{2}+\alpha \cdot r\right)^2}-L
\label{eq:displacement}
\end{equation}

It should be noted that when the actuator rotates, the amount of $\alpha$ increases, and in this case $S$ loses its significance over time. However it is quite an important parameter in the beginning, and also it adds an incremental error in each turn.\\

\subsection{String Bundle Diameter}

\begin{figure}[tbph]
      \centering
      \includegraphics[trim =50 310 280 30, clip=true, width=70 mm]{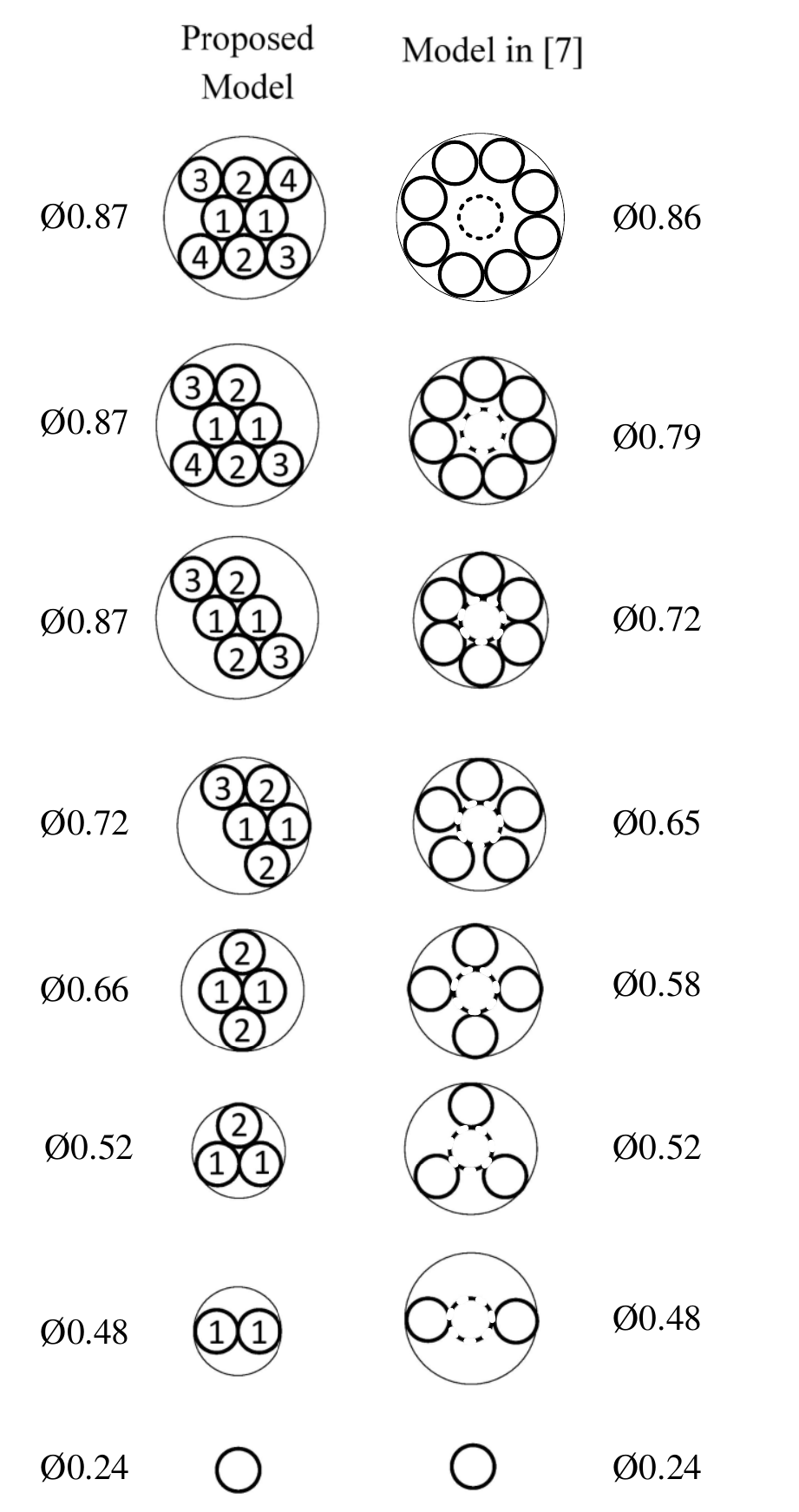}
      \caption{Difference in diameter between the existing \cite{TwistedStringActuationSystemApril2013} (right) and the proposed (left) string sections models. The numbers represent the string pairs.}
      \label{fig:stringSections}
\end{figure}

The second parameter that we studied in the above equation was the radius. Introducing the precise value of $r$ to the mathematical model is very important. The significance of $r$ in the equation increases by each rotation, since $r$ is multiplied by the increasing $\alpha$. While the radius of a single string in the twisting string system is constant, after twisting around each other, the radius of the entire string bundle is not a simple multiplication of the string radius and the number of strings. Palli et. al. suggested a model for estimating the radius of the string bundle after twisting \cite{TwistedStringActuationSystemApril2013}. \textbf{In their model (a simplifying hypothesis) some strings contribute to the total axial force: these fibers form the core.}\\

In order to find the actual diameter of the strings, when they are twisted and loaded, the strings were twisted for 5 turns with a 2 kgf load applied. Then we measured the diameter of the string bundle with a digital caliper. Each string was measured in 5 different locations and the average was calculated. The results are presented in table \ref{table:diameter}.\\

However, the experimental measurements in table \ref{table:diameter} do not match the concept presented in \cite{TwistedStringActuationSystemApril2013}. Therefore a different model is proposed here, where the center core is made of two strings, rather than a single string or an empty space, while any additional pair of strings wraps around the core, meaning that the other string pairs also wrap around the core at an equal distance from the core. In a twisted string system, one string should wrap around another string, so the first two strings that meet each other form the core. Other strings also find their pair milliseconds after the first core is formed, so they should wrap around the first formed core. Figure \ref{fig:stringSections} compares both models. As can be seen, the  model proposed in this work, shows a much better correlation with the measured diameter in table \ref{table:diameter} than the model proposed in \cite{TwistedStringActuationSystemApril2013}.\\

To describe how this format of strings is formed, we should consider the time that each pair of string reaches the core pair. For instance considering the 4 string case, when the second pair approaches the core, the strings try to place themselves around the core and as near as possible to the center. Now, to consider the case of 6 strings, we should first consider a 4 string bundle is already formed and the fifth and sixth string place themselves in a way to be as close as possible to the center (since the twisting force slides them to the center).\\

It is also important to consider why the strings organize themselves in pairs. We believe the reason for this is that the axial load is unevenly distributed along the string pairs. The two string system, in reality, has only one string, since the two ends of the string are tied to the plastic part that connects the system to the weights.
So the load is equal on the pair of strings of the two string system. When more strings are added, they are connected in the same way and because of this, each string in the pairs is equally loaded. However, different pairs may have different loads, because they are manually assembled and the length of each pair can be slightly different. Therefore strings wrap around the center at different times. For example, the pair with the higher axial load, is more stretched and reaches the center sooner and thus it forms the core.
Another reason for using only even number of strings is to apply symmetrical loads to the spinning shaft of the motor. In a symmetrical system with an even number of strings that are at the same tension, the lateral forces cancel out each other. However having an odd number of strings in this system, results in application of undesired lateral forces to the motor shaft.\\

\begin{table}[tbph]
\caption{String bundle diameter}
\label{table:diameter}
\renewcommand{\arraystretch}{1.2}
\centering
\begin{tabular}{|c|c|c|c|c|}
\hline
Nr. & Average & Standard & Proposed & Existing model\\
of & measured & deviation & model & from \cite{TwistedStringActuationSystemApril2013}\\
strings & diameter [mm] & [mm] & error [mm] & error [mm]\\
\hline
1 & 0.24 & 0.02 & 0.00 & 0.00\\
\hline
2 & 0.47 & 0.03 & 0.01 & 0.01\\
\hline
3 & 0.61 & 0.03 & -0.09 & -0.09\\
\hline
4 & 0.70 & 0.02 & -0.04 & -0.12\\
\hline
5 & 0.84 & 0.05 & -0.12 & -0.19\\
\hline
6 & 0.86 & 0.02 & 0.01 & -0.14\\
\hline
7 & 0.93 & 0.02 & -0.06 & -0.14\\
\hline
8 & 0.99 & 0.03 & -0.12 & -0.13\\
\hline
\end{tabular}
\end{table}

\subsection{Variable vs. Constant Radius Model}

\begin{figure*}[!t]
      \centering
      \includegraphics[trim =0 20 20 0, clip=true, width=180 mm]{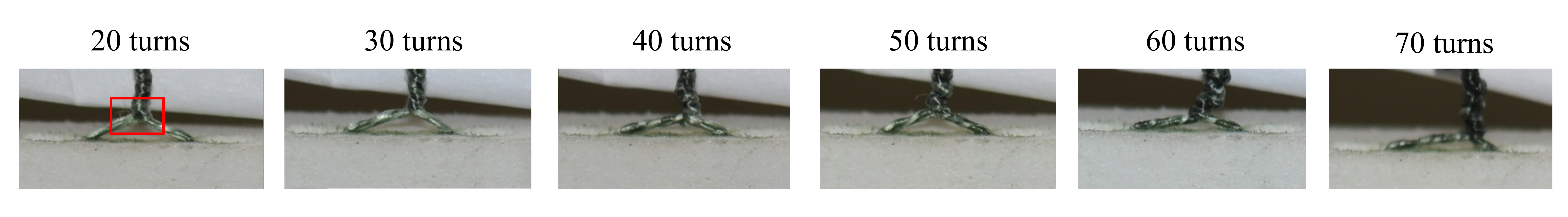}	   
      \caption{Zoom of strings under twisting close to the separator. The red rectangle highlights the zone where the strings twist when they come out of the separator. In that zone the string bundle diameter remains constant for the entire range of overtwisting. }
      \label{fig:StringsZoom}
\end{figure*}

Another part of the mathematical model that we studied was the variable radius model. The model for twisting string systems which is proposed in \cite{TwistedStringExoApril2013, TwistedStringExoNovember2013, TwistedStringExoOctober2012} presents a variable radius model to improve the correlation between the experimental results and the theory. The variable radius model considers that the radius of the string bundle will increase with twisting, because the length of the twisting zone is constant and the volume of string that enter the twisting zone keeps increasing. So it is logical that with twisting, the radius of the string bundle increases accordingly. In the variable radius model, the radius varies according to the following equation:

\begin{equation}
r_{var}=r_0 \cdot \sqrt{\frac{L+x}{L}}
\label{eq:variableRadius}
\end{equation}
Where $r_0$ is the initial radius of the bundle.\\

\begin{figure}[!h]
      \centering
       \includegraphics[trim =140 130 140 120, clip=true, width=95 mm]{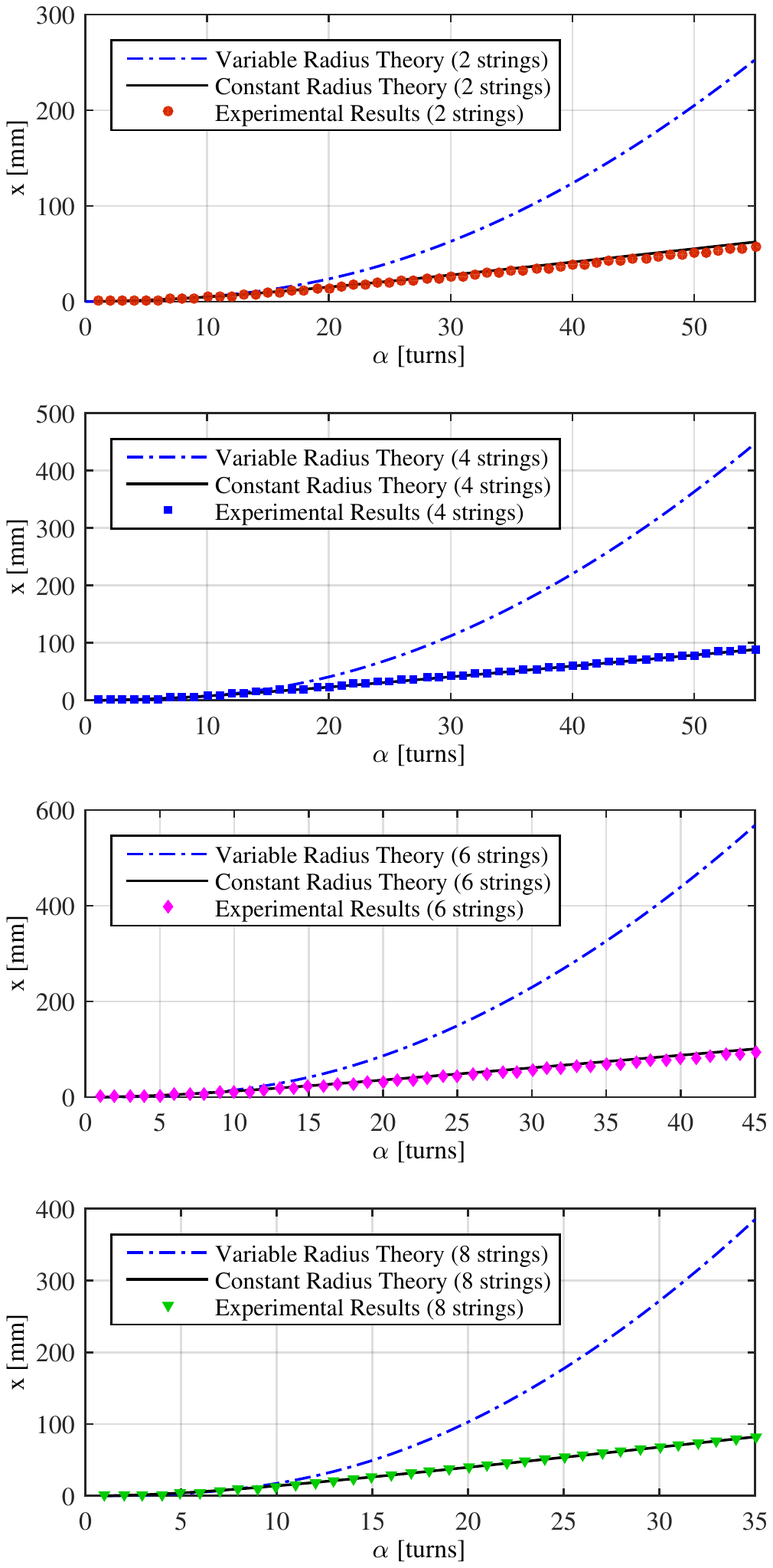}
       \caption{\textbf{Comparison between the mathematical models and the experimental results, with $x$ being the linear displacement and $\alpha$ the number of turns at the output shaft of the gearmotor.}}
      \label{fig:graphMathModels}
\end{figure}

In table \ref{table:RMSE} the two models are compared with the experimental results by using the root-mean-square deviation (RMSE). The constant radius model has the smaller RMSE values, thus it is more accurate than the variable radius model. Figure \ref{fig:graphMathModels} shows the comparison between the experimental results and variable radius model.  As can be seen, the variable radius model fails to predict the displacement in the overtwisting phase and can only match the experimental result in the first phase. The reason for this is that in the first phase the increase in the radius is uniformly distributed along the bundle in the twisting zone and when more string enters \textbf{into} the zone, the value of $r$ parameter in figure \ref{fig:stringsSchematics} and thus in equation \ref{eq:displacement} increases. As can be observed in Figure \ref{fig:strings}, on the second phase, overtwisting happens locally in different zones of the string. This means that several knots are formed in different places until the bundle is filled with those knots. Thus when overtwisting happens, the radius of the string bundle stops to increase uniformly along the length of the twisting zone and the variable radius model is no longer valid.\\ 

\begin{table}[!h]
\caption{RMSE comparison between the constant and the variable radius model}
\label{table:RMSE}
\renewcommand{\arraystretch}{1.2}
\centering
\begin{tabular}{|c|c|c|}
\hline
Nr. & RMSE Constant & RMSE Variable\\
of strings & radius model [mm] & radius model [mm]\\
\hline
2 & 2.73 & 79.29\\
\hline
4 & 5.86 & 155.88\\
\hline
6 & 4.42 & 219.63\\
\hline
8 & 3.23 & 142.78\\
\hline
\end{tabular}
\end{table}

Moreover, in the Figure \ref{fig:StringsZoom} it is possible to see that, in the area where the strings twist when they come out of the separator, the helix angle, $\beta$, and the radius, $r$, remains constant through the entire overtwisting phase. Since this is the place where the two strings actually twist around each other and the radius is constant in this part of the twisting zone, then this may be the reason for the constant radius model to work over the entire range of contraction, even if the radius increases in the rest of the twisting zone.\\

Figure \ref{fig:graphMathModels} also compares the mathematical model presented in this section with the experimental results. This mathematical model takes into the account the distance between the holes in the separator, $S$, the corrected bundle diameter model and the constant radius theory. The model matches very well the experimental results in all cases of 2, 4, 6 and 8 strings.\\

\section{Life Cycle}
\label{sec:LifeCycle}

An important aspect of the twisting string system that requires further analysis is the life cycle of the strings. \textbf{Here, we specifically intend to evaluate the life cycle of this system operating in the over-twist region and compare it to conventional systems (low contraction).}

\subsection{Experimental Setup}
\textbf{In the experiments, we study the effect of 2 different factors, namely applied load and the number of shaft rotations per cycle, on string behavior.} Three different loads (2, 3 and 5 kgf) were tested using the twisted string system with 2 strings, until the strings break. The cycle consists of twisting the strings for a fixed number of turns (20, 30, 40, 55) and then untwist the strings back to the original position. The 20 turns correspond to the limit where the overtwist phase will begin, and corresponds to the maximum contraction without overtwisting the strings. While the 55 turns are the limit where overtwisting is possible without untwisting problems. The objective is to compare the effect of different loads and the overtwist phase on the life the strings.\\

\textbf{The experimental setup consists of a gearmotor, an optical encoder and a motor driver that outputs the shaft position and measures/reports the motor current for system failure detection.} A STM32F4 microcontroller is programmed to conduct the cycles. After each cycle a short rest time of a couple of seconds is \textbf{programmed to reduce the motor temperature}. A log file is automatically created with the duration of the test, the number of turns and the current supplied to the motor. Thus, there is a record of the amount of twisting that the strings endured and the exact moment where they broke (by analyzing the current.)

\subsection{Life Cycle Analysis}
\begin{figure}[!h]
      \centering
      \includegraphics[trim =120 120 120 130, clip=true, width=90 mm]{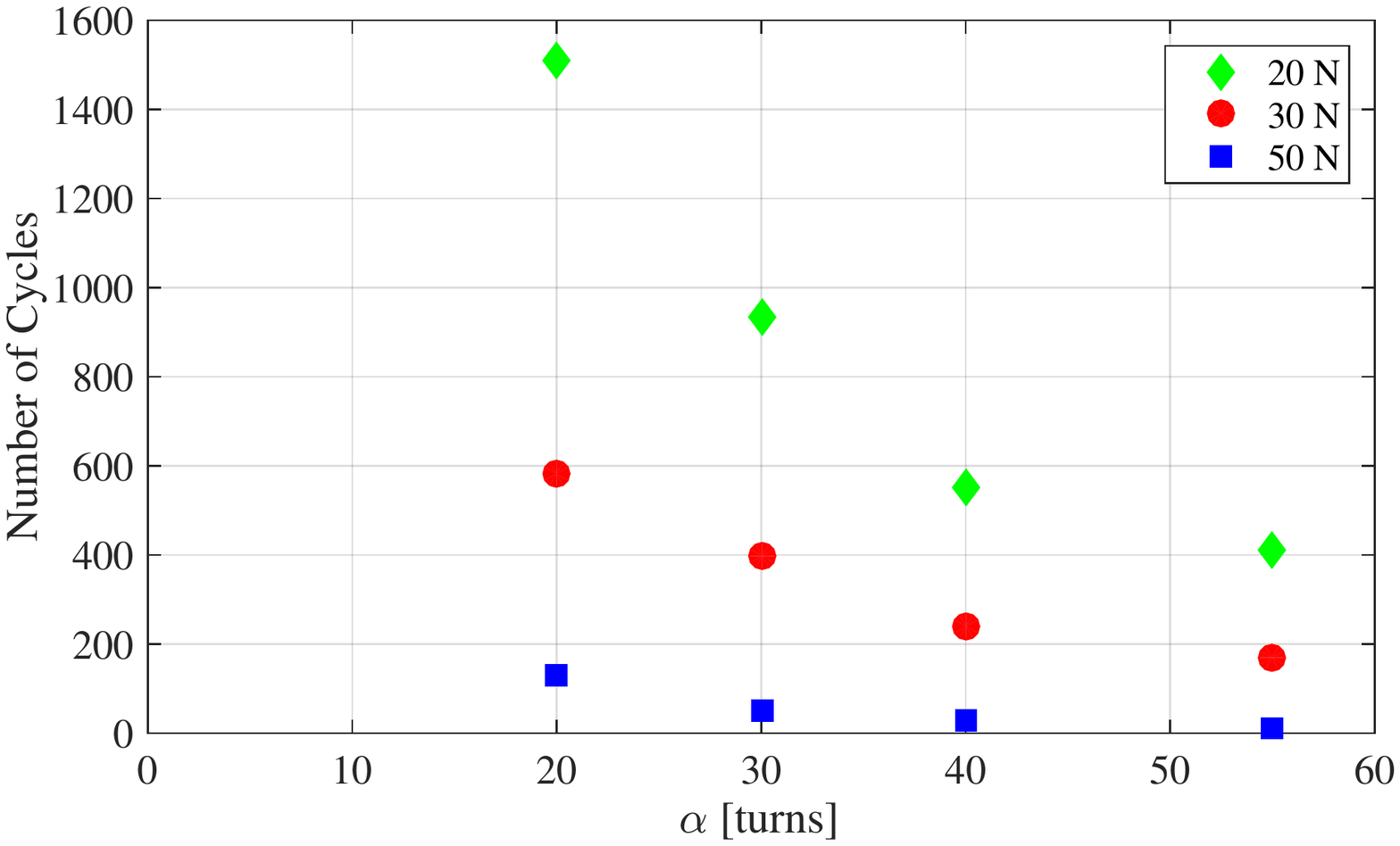}	   
      \caption{Number of cycles endured by the strings, for different number of turns, $\alpha$, and different loads.}
      \label{fig:graphLifeCycle}
\end{figure}

\begin{figure}[!h]
      \centering
      \includegraphics[trim =120 120 120 130, clip=true, width=90 mm]{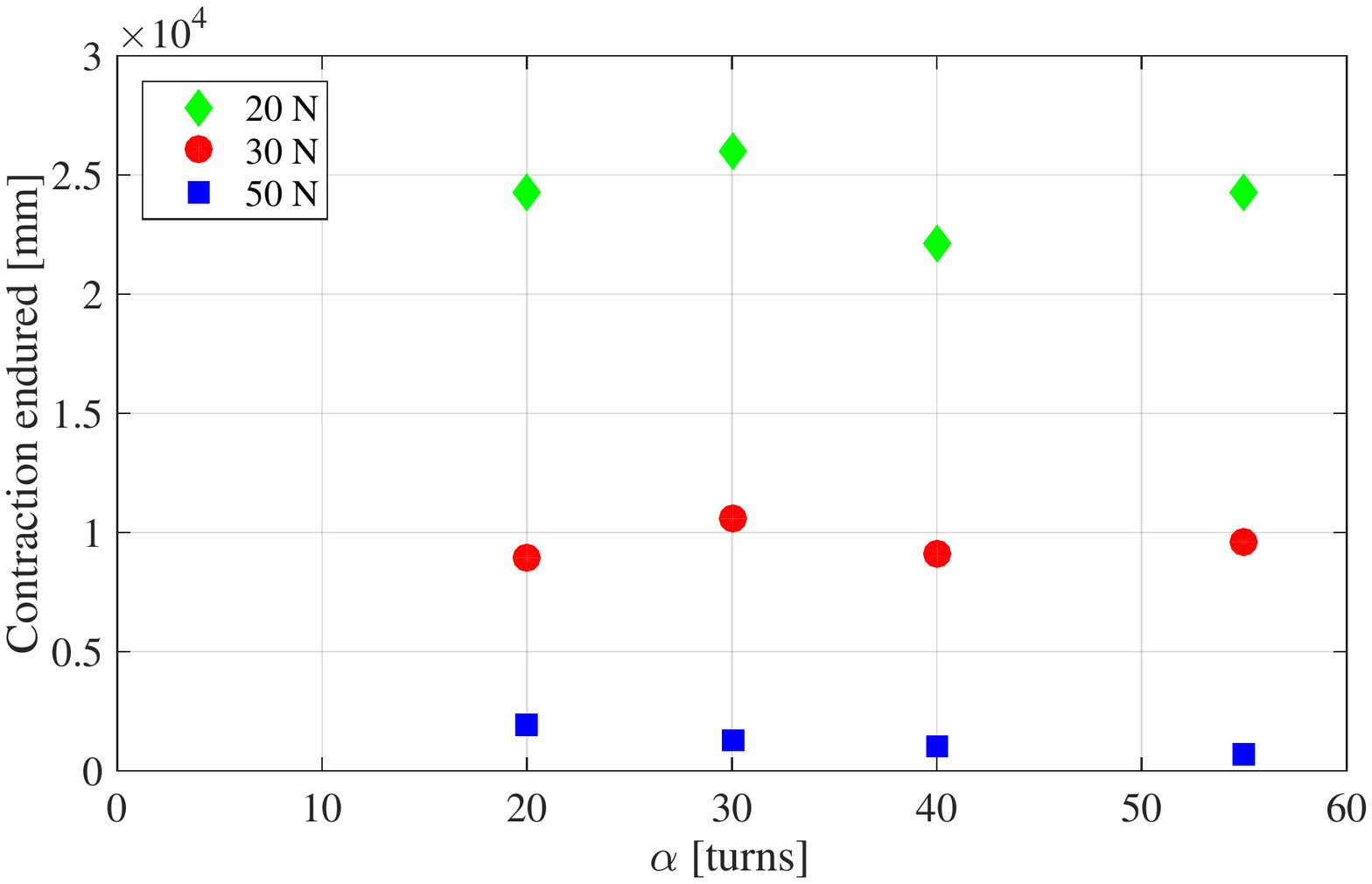}	   
      \caption{Amount of contraction endured by the strings, for different number of turns, $\alpha$, and different loads.}
      \label{fig:graphLifeCycleContraction}
\end{figure}

Table \ref{table:lifeCycle} and Figures \ref{fig:graphLifeCycle} and \ref{fig:graphLifeCycleContraction} show the results of the life cycle test.
It should be noted that the contraction endured is only half of the actual contraction the strings actually suffered. For example, in the case of 2 kgf and 30 turns, in one cycle the strings contract from 0 mm at 0 turns to 27.76 mm at 30 turns, and then they unwind back to 0 turns and 0 mm of contraction. \textbf{Table \ref{table:lifeCycle} shows the life cycle of the strings under overtwist.}\\

\begin{table}[tbph]
\caption{Life cycle of the strings under overtwist.}
\label{table:lifeCycle}
\renewcommand{\arraystretch}{1.2}
\centering
\begin{tabular}{|c|c|c|c|c|}
\hline
Applied  & Nr. of & Cycles & Contraction achieved & Contraction\\
load [kgf] & turns & endured & on each cycle [mm]& endured$^1$ [mm]\\
\hline
2 & 20 & 1512 & 16.06 & 24283\\
\hline
2 & 30 & 936 & 27.76 & 25983\\
\hline
2 & 40 & 550 & 40.22 & 22121\\
\hline
2 & 55 & 411 & 59.02 & 24257\\
\hline
\hline
3 & 20 & 581 & 15.43 & 8965\\
\hline
3 & 30 & 396 & 26.79 & 10609\\
\hline
3 & 40 & 240 & 37.64 & 9132\\
\hline
3 & 55 & 170 & 56.36 & 9581\\
\hline
\hline
5 & 20 & 130 & 14.60 & 1898\\
\hline
5 & 30 & 50 & 25.44 & 1272\\
\hline
5 & 40 & 28 & 36.08 & 1010\\
\hline
5 & 55 & 13 & 54.97 & 751\\
\hline
\multicolumn{5}{l}{$^1$ Corresponds to the number of cycles endured multiplied by the} \\
\multicolumn{5}{l}{contraction achieved on each cycle.} \\
\end{tabular}
\end{table}

As expected, the life of the strings is dependent on the applied load. The maximum contraction endured by the strings is approximately constant for strings under small loads (2 kgf and 3 kgf). However, this implies that the number of cycles decreases with the increase of the number of rotations per cycle. Nevertheless, this means that under moderate loads overtwisting does not accelerate the string failure.
As an example, under a 2 kgf force, the overall amount of contraction endured by the string in the experiment with 20 turns per cycle is very close to the experiment with 55 rotations per cycle. However, this means that the number of cycles that the latter system can undergo is smaller than the former.
Under heavier loads such as 50 N the overtwist phase seems to decrease the life of the strings, as can be observed in table \ref{table:lifeCycle}.\\

The life cycle of the strings depends on several factors, such as the amount of load, the experimental setup, number of strings, number of cycles, type of the strings, etc. This is still an open problem for the twisted string system which should be further analyzed and addressed.

\section{CONCLUSIONS}

In this article we presented a two phase twisting string system that offers a contraction percentage of up to 81\%, which is significantly higher than all previously studied systems. Furthermore, we adapted the existing mathematical model for the twisted string systems to form a mathematical model for the overtwisting case. This model is able to predict the linear displacement of the system for the entire range of contraction. We also presented a model for accurate prediction of the string bundle diameter for multi string systems that is essential for precise prediction of the system displacement. In addition, we showed that in the mathematical model for the overtwist case, a constant radius theory should be used instead of the variable radius theory. Finally a life cycle test was performed for 3 different loads. For each load 4 different rotation numbers per cycle were experimented. Results showed that the life of the string is highly dependent to the applied load. For smaller loads, the overall system displacement before the string failure is almost constant for all different number of rotations per cycle.  Also results show that for the higher loads, overtwisting accelerated the strings failure.

\section{ACKNOWLEDGMENT}
This research work was partially supported by the
Portuguese Foundation of Science and Technology,
SFRH/BPD/70557/2010 \& PEst-C/EEI/UI0048/2013

\bibliographystyle{IEEEtran}
\bibliography{references}

\begin{thebibliography}{10}
\providecommand{\url}[1]{#1}
\csname url@samestyle\endcsname
\providecommand{\newblock}{\relax}
\providecommand{\bibinfo}[2]{#2}
\providecommand{\BIBentrySTDinterwordspacing}{\spaceskip=0pt\relax}
\providecommand{\BIBentryALTinterwordstretchfactor}{4}
\providecommand{\BIBentryALTinterwordspacing}{\spaceskip=\fontdimen2\font plus
\BIBentryALTinterwordstretchfactor\fontdimen3\font minus
  \fontdimen4\font\relax}
\providecommand{\BIBforeignlanguage}[2]{{%
\expandafter\ifx\csname l@#1\endcsname\relax
\typeout{** WARNING: IEEEtran.bst: No hyphenation pattern has been}%
\typeout{** loaded for the language `#1'. Using the pattern for}%
\typeout{** the default language instead.}%
\else
\language=\csname l@#1\endcsname
\fi
#2}}
\providecommand{\BIBdecl}{\relax}
\BIBdecl

\bibitem{Strand-MuscleOctober2007}
\emph{Climbing and Walking Robots, Towards New Applications}.\hskip 1em plus
  0.5em minus 0.4em\relax Itech Education and Publishing, 2007, ch. Complex and
  Flexible Robot Motions by Strand-Muscle Actuators.

\bibitem{Strand-Muscle2004}
M.~Suzuki and A.~Ichikawa, ``Toward springy robot walk using strand-muscle
  actuators,'' in \emph{Proc. 7th Int. Conf. Climbing and Walking Robots},
  Madrid, 2004, pp. 467--474.

\bibitem{Strand-Muscle2007}
M.~Suzuki, T.~Mayahara, and A.~Ishizaka, ``Redundant muscle coordination of a
  multi-dof robot joint by online optimization,'' in \emph{Advanced intelligent
  mechatronics, 2007 IEEE/ASME international conference on}, Sept 2007, pp.
  1--6.

\bibitem{DEXMARTJuly2013}
G.~Palli, S.~Pirozzi, C.~Natale, G.~De~Maria, and C.~Melchiorri, ``Mechatronic
  design of innovative robot hands: Integration and control issues,'' in
  \emph{Advanced Intelligent Mechatronics (AIM), 2013 IEEE/ASME International
  Conference on}, July 2013, pp. 1755--1760.

\bibitem{TwistDrive2010}
I.~Godler and T.~Sonoda, ``A five fingered robotic hand prototype by using
  twist drive,'' in \emph{Robotics (ISR), 2010 41st International Symposium on
  and 2010 6th German Conference on Robotics (ROBOTIK)}, June 2010, pp. 1--6.

\bibitem{TwistDriveNov2011}
------, ``Performance evaluation of twisted strings driven robotic finger,'' in
  \emph{Ubiquitous Robots and Ambient Intelligence (URAI), 2011 8th
  International Conference on}, Nov 2011, pp. 542--547.

\bibitem{TwistedStringActuationSystemApril2013}
G.~Palli, C.~Natale, C.~May, C.~Melchiorri, and T.~Wurtz, ``Modeling and
  control of the twisted string actuation system,'' \emph{Mechatronics,
  IEEE/ASME Transactions on}, vol.~18, no.~2, pp. 664--673, April 2013.

\bibitem{TwistedStringExoOctober2012}
D.~Popov, I.~Gaponov, and J.-H. Ryu, ``A study on twisted string actuation
  systems: Mathematical model and its experimental evaluation,'' in
  \emph{Intelligent Robots and Systems (IROS), 2012 IEEE/RSJ International
  Conference on}, Oct 2012, pp. 1245--1250.

\bibitem{TwistedStringExoApril2013}
D.~Popov, I.~Gaponov, and J.~Ryu, ``A preliminary study on a twisted
  strings-based elbow exoskeleton,'' in \emph{World Haptics Conference (WHC),
  2013}, April 2013, pp. 479--484.

\bibitem{TwistedStringExoNovember2013}
D.~Popov, I.~Gaponov, and J.-H. Ryu, ``Bidirectional elbow exoskeleton based on
  twisted-string actuators,'' in \emph{Intelligent Robots and Systems (IROS),
  2013 IEEE/RSJ International Conference on}, Nov 2013, pp. 5853--5858.

\bibitem{TwistedStringExoAug2014}
I.~Gaponov, D.~Popov, and J.-H. Ryu, ``Twisted string actuation systems: A
  study of the mathematical model and a comparison of twisted strings,''
  \emph{Mechatronics, IEEE/ASME Transactions on}, vol.~19, no.~4, pp.
  1331--1342, Aug 2014.

\end{thebibliography}

%








\end{document}